%% file: main.tex
\documentclass[runningheads]{llncs}

\usepackage[mobile]{eccv}

\usepackage{eccvabbrv}

\usepackage{graphicx}
\usepackage{booktabs}

\usepackage[accsupp]{axessibility}  

\input{preamble}

\usepackage{hyperref}

\usepackage{orcidlink}

\author{
Dingrui Wang\inst{1,2,*}
\quad
Zhihao Liang\inst{1,*} 
\quad
Hongyuan Ye\inst{1,*} 
\quad
Zhexiao Sun\inst{1,*} 
\quad
Zhaowei Lu\inst{1}
\quad
Yuchen Zhang\inst{1}
\quad
Yuyu Zhao\inst{1}
\quad
Yuan Gao\inst{1}
\quad
Marvin Seegert\inst{1}
\quad
Finn Schäfer\inst{1}  
\quad
Haotong Qin\inst{3}
\quad
Wei Li\inst{4}
\quad
Luigi Palmieri\inst{2} 
\quad
Felix Jahncke\inst{1}
\quad
Mattia Piccinini\inst{1}
\quad
Johannes Betz\inst{1} 
}

\authorrunning{Wang et al.}

\institute{
\centering
$^1$TUM \hspace{1.5em} 
$^2$Bosch AI Center\hspace{1.5em}
$^3$ETH\hspace{1.5em} 
$^4$NJU  \\ \url{https://target-bench.github.io/}
}

\begin{document}

\title{\raisebox{-1.0ex}{\includegraphics[height=2.0\baselineskip]{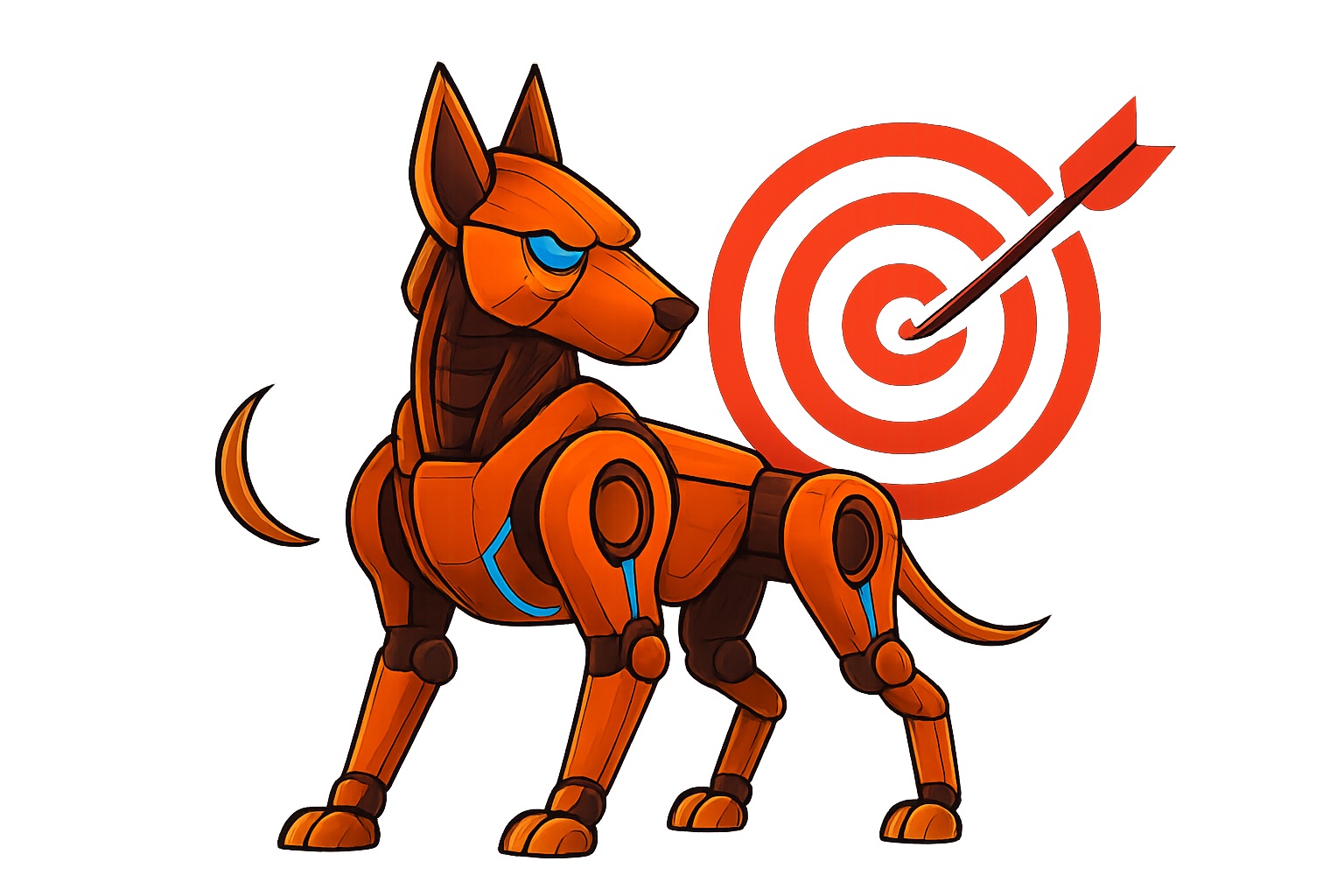}} {\LARGE \textbf{Target-Bench}}: \\Can Video World Models Achieve \\Mapless Path Planning with Semantic Targets?
}
\titlerunning{Target-Bench}





{
\maketitle
\begin{center}
\includegraphics[width=0.95\linewidth]{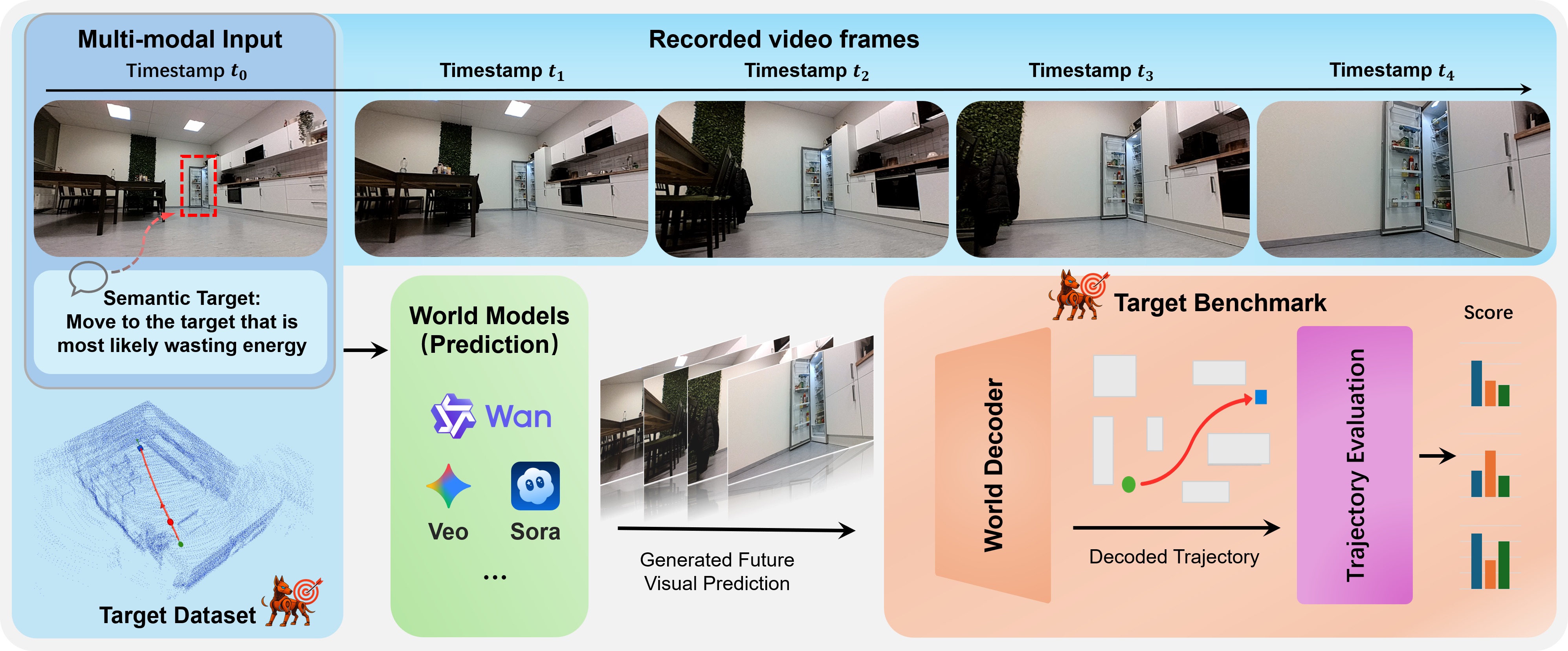}
\end{center}
\captionsetup{type=figure}
\captionof{figure}{%
A Video World Model’s planning ability is evaluated by our \textbf{benchmark}. We provide a diverse \textbf{dataset} containing annotated semantic goal, task description and motion reference to establish this novel benchmark. 
The generated future visual predictions are processed by our \textbf{world-decoder}, and the results are then evaluated with meticulously designed metrics that focus on tendency rather than trajectory overlap. 
}\label{fig:teaser}
}

\blfootnote{*equal contribution}

\input{sec/0_abstract}
\input{sec/1_intro}

\input{sec/2_rw}
\input{sec/3_method}

\input{sec/4_exp}
\input{sec/5_ds}
\input{sec/6_ccl}
\newpage
{
    \small
    \bibliographystyle{splncs04}
    \bibliography{main.bib}
}


\end{document}

%% file: preamble.tex
\usepackage{booktabs}
\usepackage{multirow}
\usepackage{graphicx}
\usepackage{array}
\usepackage{gensymb}  
\usepackage{fancyvrb}
\usepackage{lipsum}
\usepackage{xspace}
\usepackage{pifont}
\usepackage{amsmath,amssymb,amsfonts}
\usepackage{algorithm}
\usepackage{algpseudocode}
\usepackage{setspace}
\usepackage{makecell}
\usepackage{subcaption}
\usepackage{wasysym}
\usepackage{tabularx}
\usepackage[acronym, nomain]{glossaries}
\makeglossaries

\definecolor{cvprblue}{rgb}{0.21,0.49,0.74}
\usepackage[pagebackref,breaklinks,colorlinks,allcolors=cvprblue]{hyperref}

\newcommand{\cmark}{\ding{51}} 
\newcommand{\xmark}{\ding{55}} 

\newcommand{\fadedtext}[1]{\textcolor{gray}{#1}}

\makeatletter
\renewcommand{\paragraph}{%
    \@startsection{paragraph}{4}%
    {\z@}{-0.5em}{-0.5em}%
    {\normalfont\normalsize\bfseries}%
}
\makeatother

\newacronym{vlm}{VLM}{Vision Language Model}
\newacronym{slam}{SLAM}{Simultaneous Localization and Mapping}
\newacronym{imu}{IMU}{Inertial Measurement Unit}
\newacronym{vggt}{VGGT}{Visual Geometry Grounded Transformer}
\newacronym{ekf}{EKF}{Extended Kalman Filter}
\newacronym{ade}{ADE}{Average Displacement Error}
\newacronym{fde}{FDE}{Final Displacement Error}
\newacronym{mr}{MR}{Miss Rate}
\newacronym{se}{SE}{Soft Endpoint}
\newacronym{ac}{AC}{Approach Consistency}
\newacronym{wm}{WM}{World Model}
\newacronym{wo}{WO}{Weighted Overall}

\newcommand\blfootnote[1]{%
  \begingroup
  \renewcommand\thefootnote{}\footnote{#1}%
  \addtocounter{footnote}{-1}%
  \endgroup
}

%% file: sec/0_abstract.tex
\begin{abstract}
While recent video world models can generate highly realistic videos, their ability to perform semantic reasoning and planning remains unclear and unquantified. We introduce \textbf{Target-Bench}, the first benchmark that enables comprehensive evaluation of video world models' semantic reasoning, spatial estimation, and planning capabilities. 
Target-Bench provides 450 robot-collected scenarios spanning 47 semantic categories, with SLAM-based trajectories serving as motion tendency references. 
Our benchmark reconstructs motion from generated videos with a metric scale recovery mechanism, enabling the evaluation of planning performance with five complementary metrics that focus on target-approaching capability and directional consistency. 
Our evaluation result shows that the best off-the-shelf model achieves only a 0.341 overall score, revealing a significant gap between realistic visual generation and semantic reasoning in current video world models. Furthermore, we demonstrate that fine-tuning process on a relatively small real-world robot dataset can significantly improve task-level planning performance.
\keywords{World models \and Semantic reasoning \and Tasking benchmark}
\end{abstract}

%% file: sec/1_intro.tex
\section{Introduction}
\label{sec:intro}
%
%
Embodied AI has been advancing rapidly, while its core challenges are becoming increasingly clear. 
Despite major improvements in robotic hardware (e.g. actuators), the embodied intelligence remains the main bottleneck limiting robots from realizing their full potential~\cite{long2025survey}.
As a result, the robotics community is seeking to develop embodied AI systems that are more robust and generalizable. 

\glspl{wm} learn to predict how the world evolves over time~\cite{ha2018worldmodels, lecun2022path}.
In early 2024, Sora~\cite{videoworldsimulators2024}, a video generation model, was the first to self-claim as a world evolution simulation model.
Recent breakthroughs in video models demonstrate remarkable reasoning abilities in visual semantics and causality~\cite{videoworldsimulators2024, wiedemer2025video, gao2025foundation}. In this regard, these video models can also be regarded as world models~\cite{ball2025genie3}. Given an initial observation (e.g., an image frame) and a condition (e.g., a text prompt), video \glspl{wm} such as Veo 3.1~\cite{veo3deepmind} and Sora 2~\cite{sora2} can generate future frames with photorealistic quality and high spatio-temporal consistency. 

Unsurprisingly, video \glspl{wm} have sparked growing interest from the robotics community~\cite{li2025robotic, zhu2025unified, pang2025learning, wang2025enhancing}. Recent approaches such as UnifoLM-WMA-0~\cite{unifolm} aim to integrate video \glspl{wm} with low-level robot control.
The underlying philosophy of these approaches is that if a model can accurately predict how the world evolves, its predictions can serve as a plan to guide the robot actions.
However, a key question remains: How accurate must these predictions be to count as useful for planning?
More broadly, \textbf{\textit{how can we quantitatively assess a video \gls{wm}’s reasoning, task-solving, and planning ability?}}
Existing evaluation frameworks focus mainly on visual fidelity and spatio-temporal consistency~\cite{li2025worldmodelbench, duan2025worldscore}, while assessing mapless path planning that require strong semantic understanding remains an elusive task~\cite{long2025survey}.
Bridging this gap requires a benchmark that can holistically evaluate video world models’ semantic reasoning beyond visual quality and spatio-temporal consistency. 

To address this challenge, we introduce \textbf{Target-Bench}. To the best of our knowledge, it is the first benchmark for evaluating video \glspl{wm} for mapless path planning toward semantic targets in unstructured real-world environments. The semantic targets are described in natural language, and many cases are designed to include implicit semantic meanings (Fig.~\ref{fig:teaser}). Our contributions are as follows:
\begin{itemize}
    \item The first systematic \textbf{benchmark} for evaluating video world models on mapless path planning with textual semantic goals, featuring a modular and replaceable world-decoder and a meticulously designed set of metrics, which makes comprehensive evaluation of video world models' semantic reasoning, spatial estimation, and planning capabilities possible.
    \item An \textbf{open-source dataset} of 450 semantic-target-oriented scenarios (112,500 frames) collected with a quadruped robot, covering 47 categories across diverse indoor and outdoor environments, with ground truth trajectories and human annotations for explicitly and implicitly described targets. 
    \item A comparative study of open-source and proprietary \glspl{wm}, including the first \textbf{fine-tuning} of an open-source model on a small real-world robot dataset for path planning, showing that data-efficient adaptation can substantially boost task-level planning performance. We open-source all code and dataset.
\end{itemize}

%% file: sec/2_rw.tex
\section{Related Work}
\paragraph{World Models.} Early approaches such as World Models~\cite{ha2018worldmodels} and Dreamer~\cite{hafner2020dreamcontrol} introduced latent state-space models, demonstrating that generative prediction has the potential to support planning and control~\cite{gao2025foundation}.
More recently, foundation-scale methods pretrained on large video corpora have emerged: Cosmos~\cite{nvidia2025cosmos} compared diffusion and autoregressive paradigms; DIAMOND~\cite{alonso2024diamond} improved visual fidelity; V-JEPA 2~\cite{assran2025vjepa2} enhanced efficiency by forecasting in latent space.
Advances in video \glspl{wm}, including Sora 2~\cite{sora2}, Veo 3.1~\cite{veo3deepmind,wiedemer2025video}, and Wan~\cite{wan2025wan}, have further improved prompt adherence and physics-aware dynamics, enabling long-horizon predictions.
Interactive \glspl{wm} such as Genie-1~\cite{bruce2024genie1}, Genie-2~\cite{parkerholder2024genie2}, and Genie-3~\cite{ball2025genie3} provide controllable environments for use in robotics and games.
Unitree's recent UnifoLM-WMA-0~\cite{unifolm} aims to extract the reasoning capabilities of \glspl{wm} and translate them into real-world actions.
However, the ability of \glspl{wm} to perform robot task planning in real-world scenarios is largely untested.

\paragraph{Semantic Navigation Datasets.}
Navigation datasets such as R2R~\cite{anderson2018vision} and RxR~\cite{ku2020room} are primarily restricted to indoor simulation environments. 
To address real-world complexity, datasets such as MuSoHu~\cite{nguyen2023toward}, EgoWalk~\cite{akhtyamov2025egowalk}, and SANPO~\cite{waghmare2025sanpo} capture multimodal data via wearable sensors, though the latter faces heading misalignment and IMU drift. 
Others utilize human videos or robot demonstrations for language-conditioned policies (LeLaN~\cite{hirose2024lelan}) and socially compliant interaction (SCAND~\cite{karnan2022socially}, SACSoN~\cite{hirose2023sacson}). 
For broader coverage, Sekai~\cite{li2025sekai} and CityWalker~\cite{liu2025citywalker} leverage web-scale videos to achieve global diversity, although trajectories are only annotated up-to-scale. 
Crucially, as compared in Table~\ref{tab:dataset_comparison}, these datasets lack explicit navigation targets, and their annotations are descriptive rather than target-oriented.

\paragraph{Benchmarks for World Models.}
Evaluation of \glspl{wm} has transitioned from assessing visual quality and camera controllability (VBench~\cite{huang2024vbench}, WorldScore~\cite{duan2025worldscore}) and basic physics (WorldModelBench~\cite{li2025worldmodelbench}) to functional utility. 
While frameworks like World-in-World~\cite{zhang2025world} and WorldArena~\cite{shang2026worldarena} enable closed-loop testing, they are predominantly confined to simulation.
Crucially, many recent robotics-oriented suites (RBench~\cite{deng2026rethinking}, EWMBench~\cite{yue2025ewmbench}, WorldEval~\cite{li2025worldeval}) rely on subjective \gls{vlm}-based judges for success detection, which introduces inherent judge uncertainty and limits result reproducibility. Moreover, existing benchmarks typically focus on explicit semantic instruction rather than implicit semantic reasoning. A concise comparison across these dimensions is provided in Table~\ref{tab:benchmark_comparison}.

\begin{figure*}[t]
\centering
\begin{subfigure}[b]{0.65\textwidth}
    \centering
    \includegraphics[width=\linewidth]{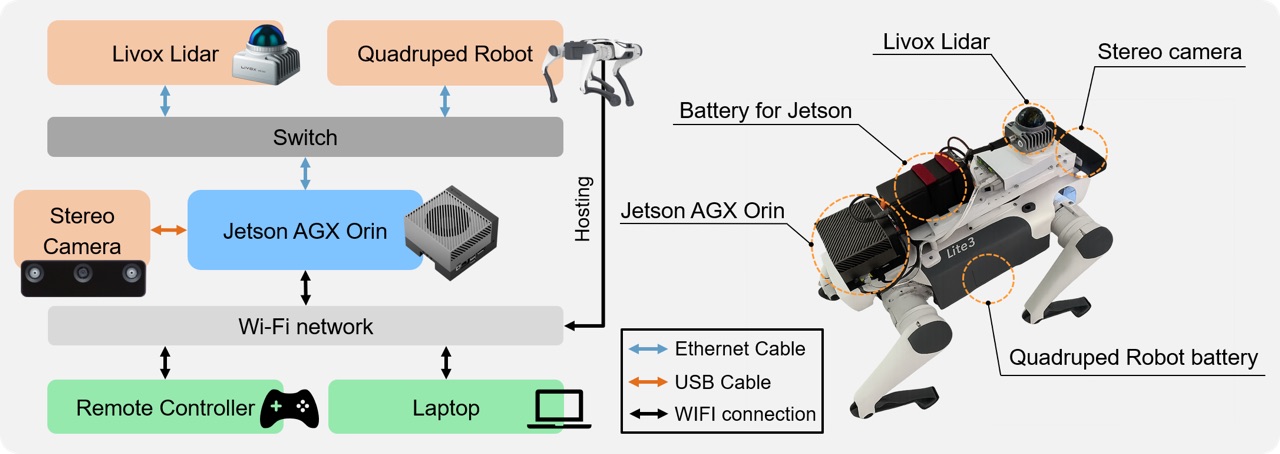}
    \caption{Quadruped robot hardware setup.}
    \label{fig:robot_setup}
\end{subfigure}
\begin{subfigure}[b]{0.33\textwidth}
    \centering
    \includegraphics[width=\linewidth]{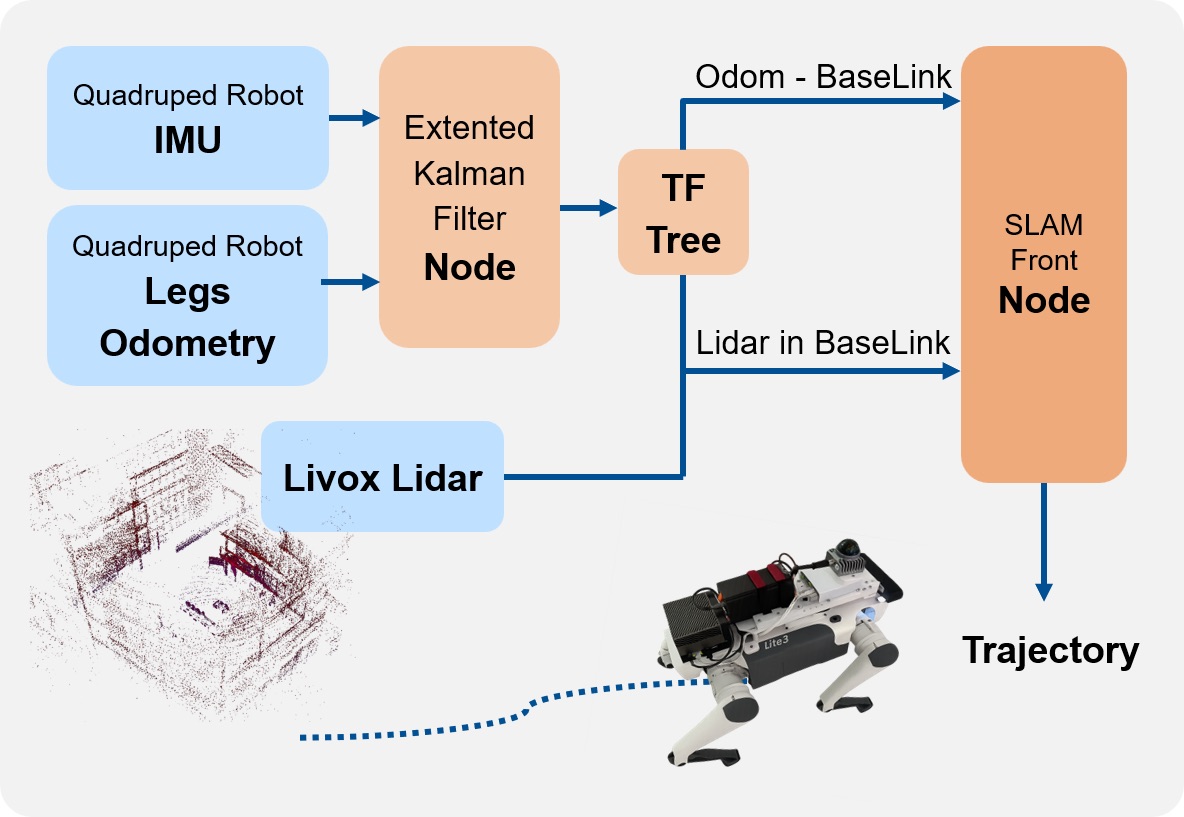}
    \caption{SLAM pipeline.}
    \label{fig:slam}
\end{subfigure}
\caption{Robot setup and SLAM pipeline.}
\label{fig:robot}
\end{figure*}

%% file: sec/3_method.tex
\begin{figure*}[t]
\centering
\begin{subfigure}[b]{0.216\textwidth}
    \centering
    \includegraphics[width=\linewidth]{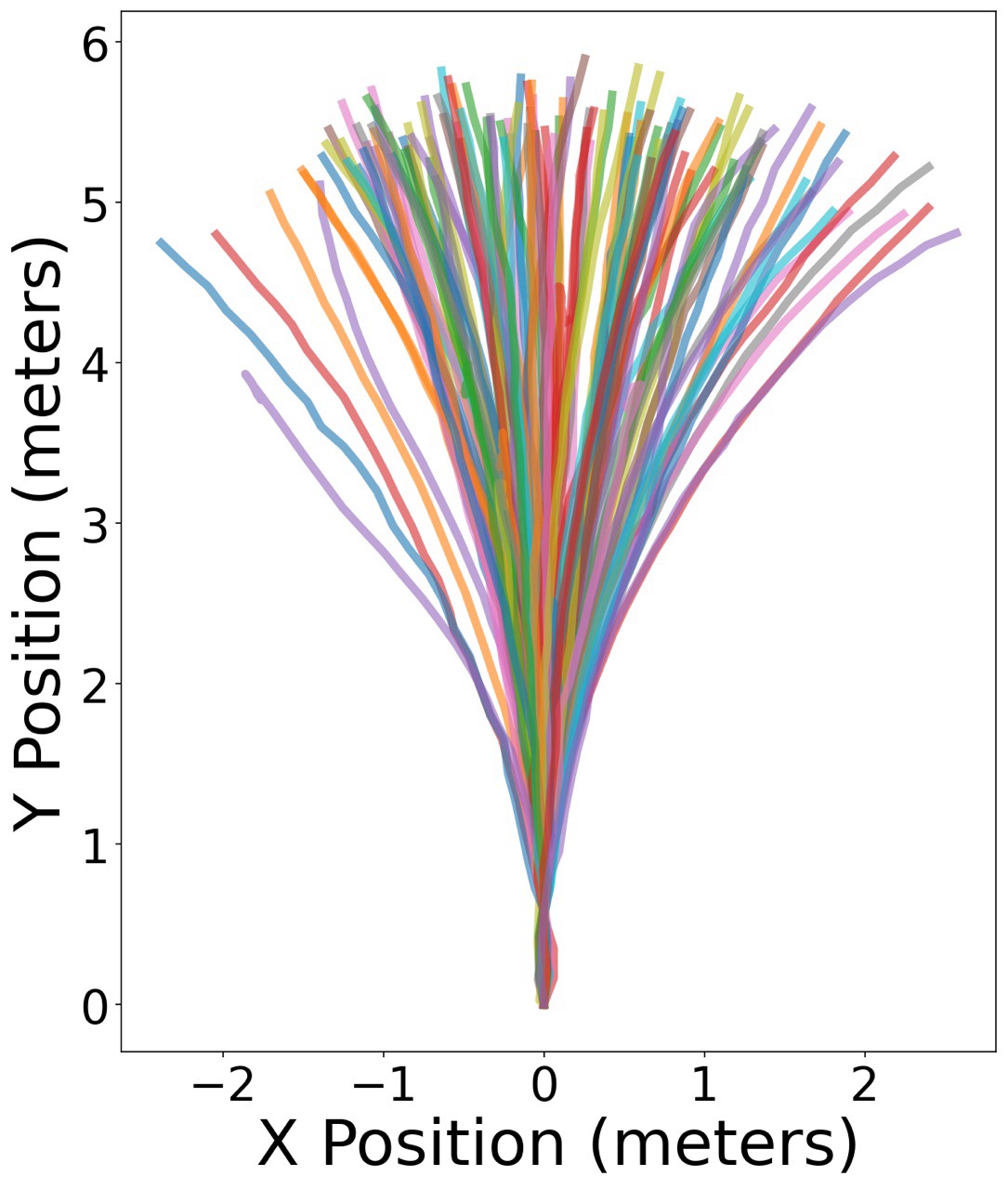}
    \caption{Trajectories.}
    \label{fig:all_traj}
\end{subfigure}
\hfill
\begin{subfigure}[b]{0.32\textwidth}
    \centering
    \includegraphics[width=\linewidth]{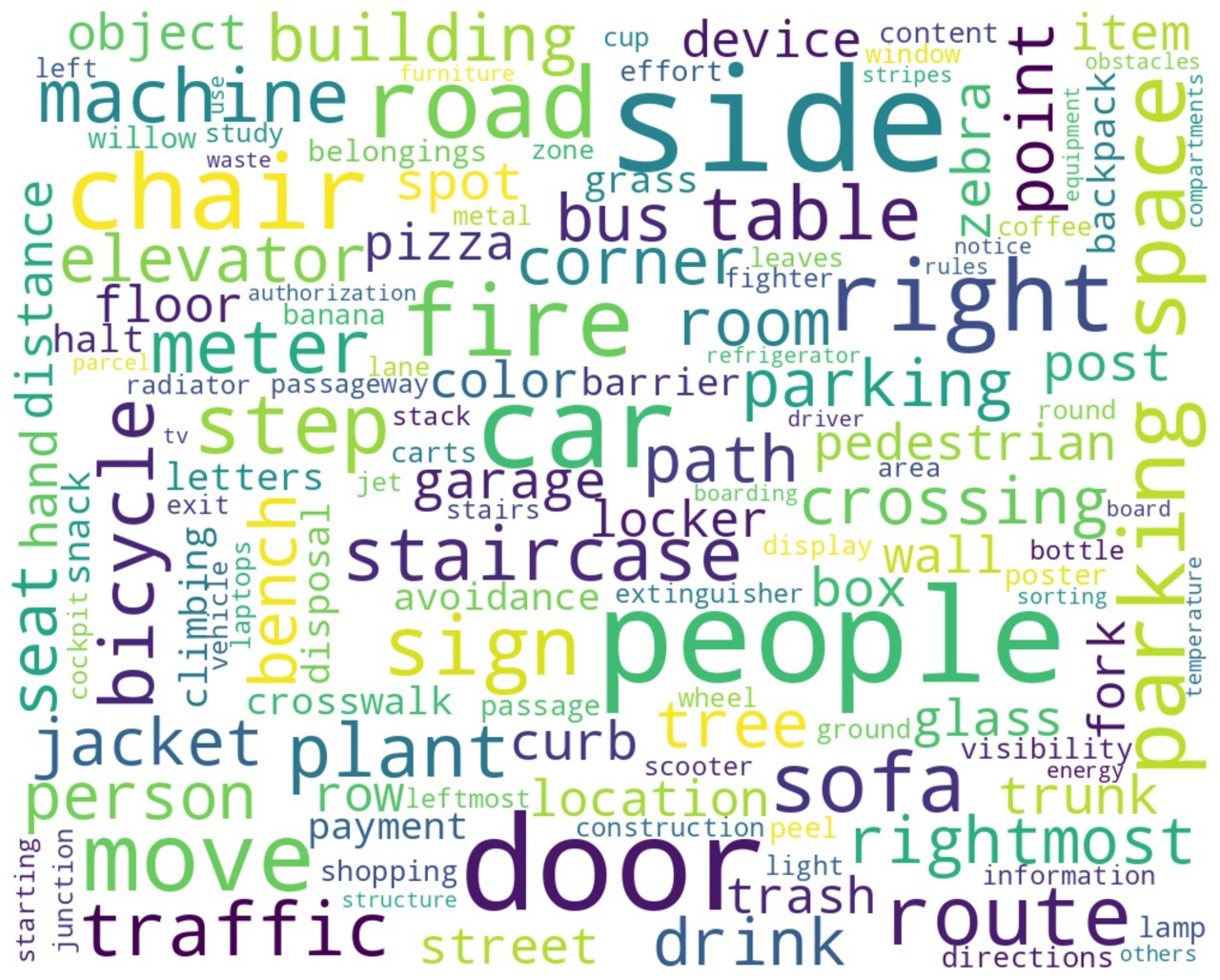}
    \caption{Caption word cloud.}
    \label{fig:wordcloud}
\end{subfigure}
\hfill
\begin{subfigure}[b]{0.447\textwidth}
    \centering
    \includegraphics[width=\linewidth]{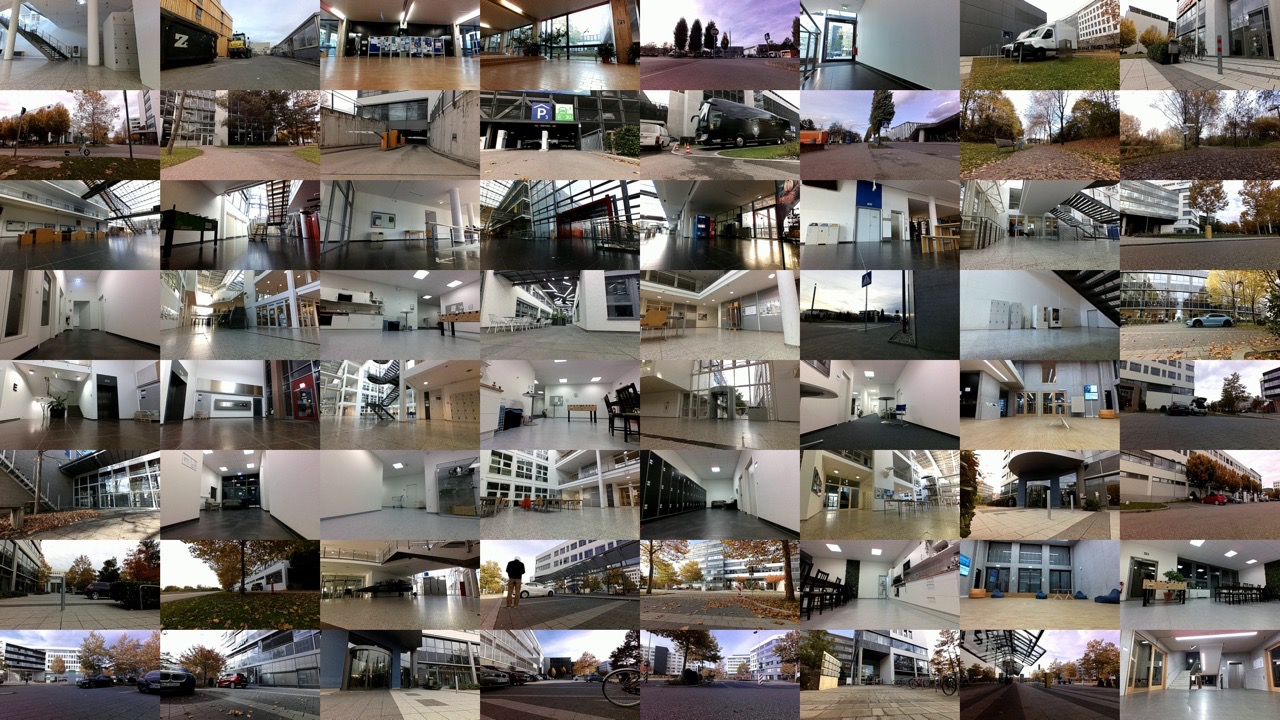}
    \caption{Data scenario environments.}
    \label{fig:env}
\end{subfigure}
\hfill
\begin{subfigure}[b]{1.0\textwidth}
    \centering
    \includegraphics[width=\linewidth]{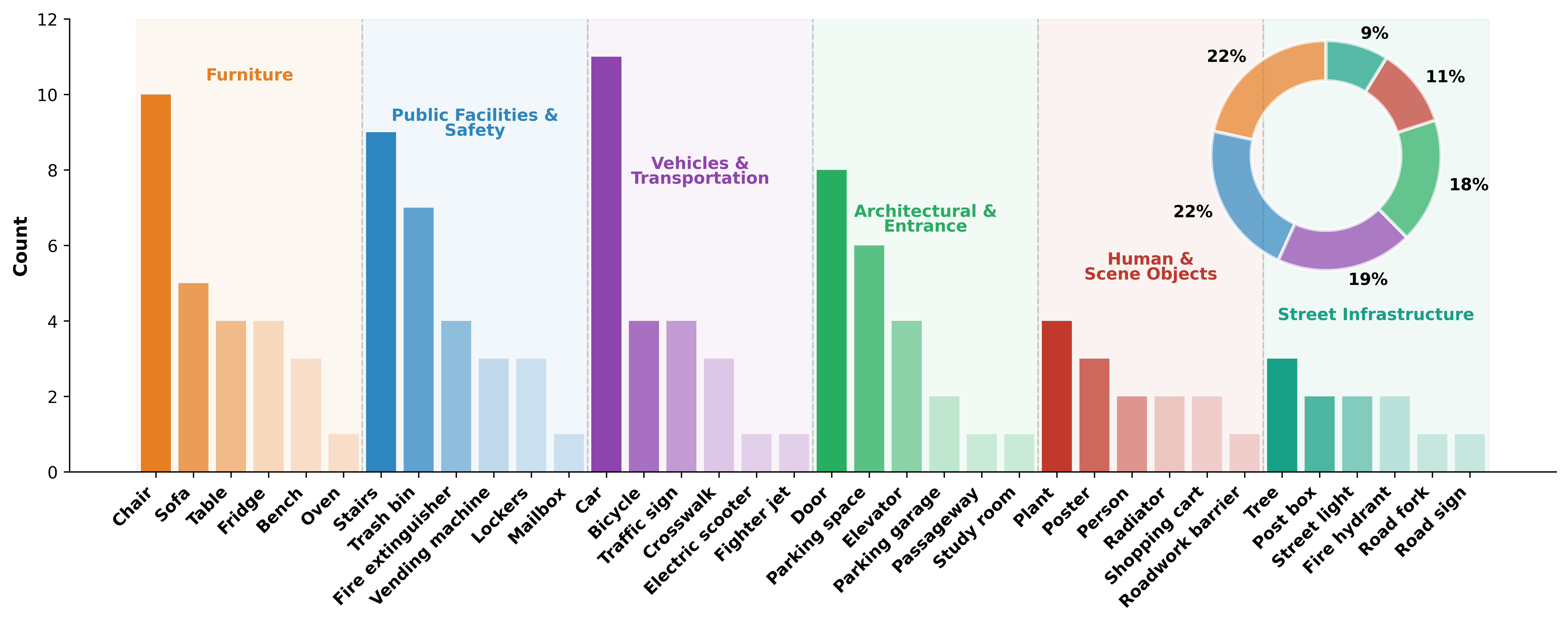}
    \caption{Semantic target categories.}
    \label{fig:category}
\end{subfigure}
\caption{Visualization of benchmark dataset structure and semantics: (a) spatial distribution of trajectories, (b) word cloud representation of caption vocabulary, (c) data collection environments, and (d) distribution of semantic target categories.}
\label{fig:dataset_viz}
\end{figure*}

\section{Target-Bench}
As shown in Fig.~\ref{fig:teaser}, Target-Bench consists of two components: a Target dataset and the Target benchmark evaluation pipeline.
The \textbf{Target dataset} (Fig.~\ref{fig:robot_setup}) is collected with a quadruped robot equipped with multi-modal sensors in diverse indoor and outdoor environments. It spans 47 semantic target categories (e.g., cars, chairs, bicycles) and includes 450 scenarios.
Our \textbf{Target benchmark} comprises two stages: world-decoder and path evaluation. The world-decoder extracts camera poses from generated video to form a path, which is then compared against the ground truth path. Path evaluation metrics focus on the proximity to the target and directional consistency, jointly assessing scenario reconstruction quality and semantic navigation capability.

\begin{table}[t]
\centering
\caption{\textbf{Comparison with existing navigation datasets.} 
Our dataset is collected via target-oriented teleoperation of a real quadruped robot in real-world environments, providing video sequences paired with semantic target prompts, point cloud maps, and metric-scale trajectories with kinematic consistency.}
\label{tab:dataset_comparison}
\scriptsize
\setlength{\tabcolsep}{3pt}
\begin{tabularx}{\linewidth}{
p{2.5cm} 
>{\centering\arraybackslash}p{2.4cm}
*{4}{>{\centering\arraybackslash}X}
}
\toprule
\textbf{Dataset} 
& \makecell[c]{\textbf{Source} \\ \textbf{Type}} 
& \makecell[c]{\textbf{Metric} \\ \textbf{Traj.}} 
& \makecell[c]{\textbf{Semantic} \\ \textbf{Target}} 
& \makecell[c]{\textbf{Point} \\ \textbf{Cloud}} 
& \makecell[c]{\textbf{Kinem.} \\ \textbf{Match}} \\
\midrule
R2R~\cite{anderson2018vision} & Simulation & \xmark & \checkmark & \xmark & \xmark \\
RxR~\cite{ku2020room} & Simulation & \xmark & \checkmark & \xmark & \xmark \\
LeLaN~\cite{hirose2024lelan} & Real-Rob./Human & \xmark & \checkmark & \xmark & $\ocircle$ \\
MuSoHu~\cite{nguyen2023toward} & Real-Human & \checkmark & \xmark & \checkmark & \xmark \\
SACSoN~\cite{hirose2023sacson} & Real-Robot & \checkmark & \xmark & \xmark & \checkmark \\
SANPO~\cite{waghmare2025sanpo} & Sim./Real & \checkmark & \xmark & \xmark & \xmark \\
Sekai~\cite{li2025sekai} & Sim./Real & \xmark & \xmark & \xmark & \xmark \\
CityWalker~\cite{liu2025citywalker} & Real-Rob./Human & $\ocircle$ & \xmark & \xmark & $\ocircle$ \\
SCAND~\cite{karnan2022socially} & Real-Robot & \checkmark & \xmark & \checkmark & \checkmark \\
EgoWalk~\cite{akhtyamov2025egowalk} & Real-Human & \checkmark & \xmark & \xmark & \xmark \\
\midrule
\textbf{Ours} 
& \textbf{Real-Robot} & \checkmark & \checkmark & \checkmark & \checkmark \\
\bottomrule
\end{tabularx}
\end{table}

\subsection{Target Dataset Setup}
\subsubsection{Quadruped Robot Platform}
\label{subsubsec:quadruped_platform}
As shown in Fig.~\ref{fig:robot_setup}, our data-collection platform is built on a DEEP Robotics Lite 3 Venture quadruped robot. It carries a Livox Mid-360 LiDAR, an OAK-D Pro W stereo RGB camera, and an NVIDIA Jetson AGX Orin for mapping and state estimation. The LiDAR connects to the robot base via an Ethernet switch, and the camera connects to the Jetson via USB. The Jetson is powered by a dedicated onboard battery. A Wi-Fi link connects the Jetson to a remote laptop for monitoring and logging, and to a handheld controller for teleoperation.
Our software stack (Fig.~\ref{fig:slam}) uses a LiDAR-centric SLAM pipeline~\cite{segalGeneralizedICP2009, kümmerleg2o2011} with multi-sensor fusion. \gls{imu} data and legged odometry are fused via an \gls{ekf} to produce a stable base-frame pose, which is broadcast through the ROS TF tree to align the LiDAR and robot frames. The LiDAR point clouds are processed by the \gls{slam} front-end with motion compensation and incremental registration guided by the fused odometry. The back-end then optimizes the trajectory and builds a global map, enabling accurate pose estimation.

\begin{figure*}[t]
\centering
\includegraphics[width=\linewidth]{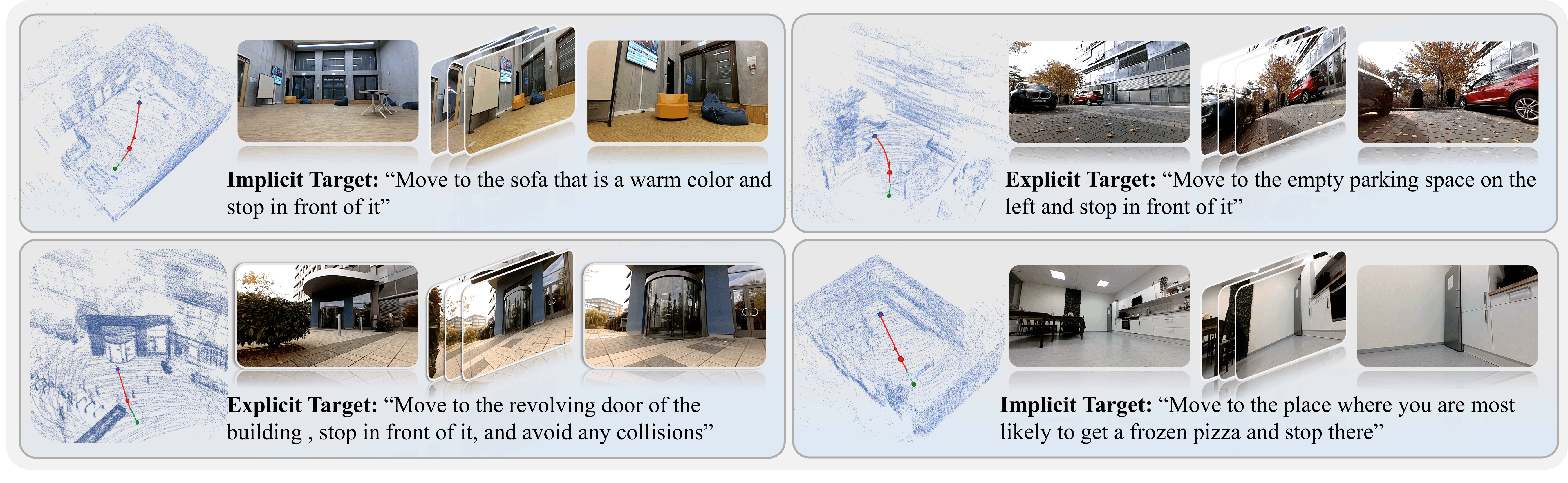}
\caption{Data Sample Visualization.}
\label{fig:scene_vis}
\end{figure*}   

\subsubsection{Semantic Target Data Collection}
The data collected in this work is to support the evaluation of video world models on mapless semantic navigation tasks. The objective is to assess whether a model can infer a plausible motion trend toward a specific semantic target in the scene. Therefore, the robot dog is always initialized with the target already present in the field of view. To guarantee a clear and unambiguous motion trend, a human operator manipulates the robot using a direct goal-approach strategy, maintaining constant forward speed and minimizing unnecessary heading oscillations. This avoids exploration and results in stable and reproducible motion patterns across different scenes.

Each processed data sample in the Target dataset consists of four components: a video sequence, a ground truth trajectory, a semantic target specified in the form of a prompt, and a point cloud map. Video frames are captured at 25Hz, yielding approximately 10 seconds of continuous observation per sample. The first 2 seconds are used for scale factor computation in Sec.~\ref{sec:world_decoder}, the latter 8 seconds are used as future frames for model inference and training.
Each prompt is a concise imperative command expressed in natural language, ensuring that the navigation outcome can be judged with respect to a single well-defined goal in the scene. The semantic targets are pre-selected by human experts to ensure diversity and relevance for navigation tasks. Each target is defined as either \emph{explicit} or \emph{implicit}. Explicit targets specify the object name directly, whereas implicit targets describe the object through its attributes or functions embedded in the prompt context (Fig.~\ref{fig:scene_vis}). Among the 450 collected samples, 232 are indoors and 218 are outdoors, 72 have implicit targets and 378 have explicit targets. 125 samples are curated as the benchmark set based on scene/target diversity and a balanced distribution of left/forward/right motion trends, while the remaining 325 samples are used exclusively as training set. Fig.~\ref{fig:dataset_viz} demonstrates the benchmark dataset structure. 

\begin{figure*}[t]
\centering
\includegraphics[width=\linewidth]{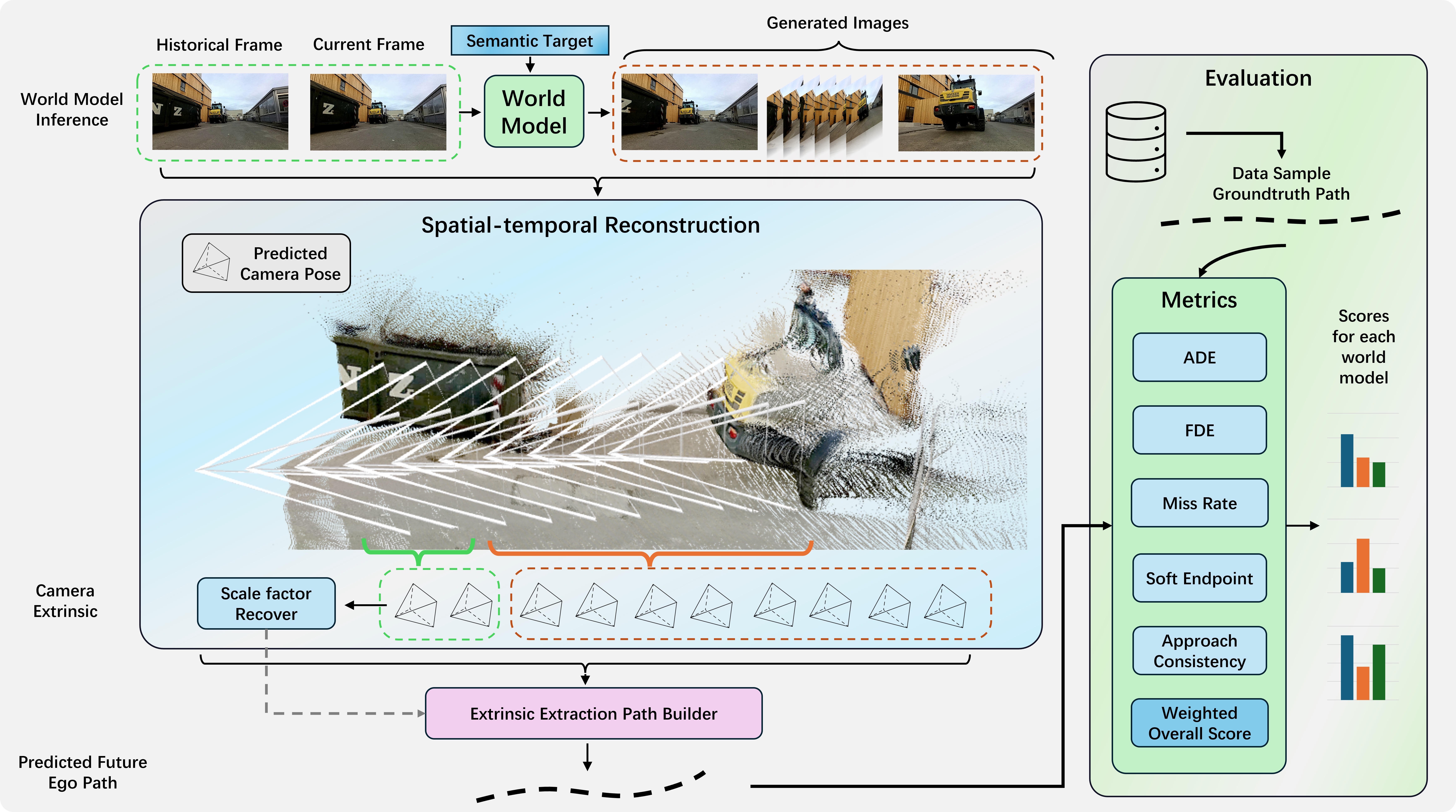}
\caption{Target Benchmark Architecture.}
\label{fig:framework}
\end{figure*} 

\subsection{Target Benchmark Architecture}
The Target benchmark (Fig.~\ref{fig:framework}) consists of two main components designed to quantify world model planning capabilities. First, the \textbf{world-decoder} extracts camera trajectories from generated videos using state-of-the-art 3D reconstruction methods. Second, the \textbf{path evaluation} module computes a score focusing on approach tendency using a comprehensive suite of metrics. 


\begin{table}[t]
\centering
\caption{\textbf{Comparison of Video World Model Benchmarks.} Target-Bench provides invariant, task-driven evaluation of world-model planning in real-world robot scenes, and includes tasks that require implicit semantic reasoning, without relying on subjective model-based scoring, thus reducing judge uncertainty.}
\label{tab:benchmark_comparison}
\scriptsize
\setlength{\tabcolsep}{2pt}
\begin{tabularx}{\linewidth}{
p{2.6cm} *{6}{>{\centering\arraybackslash}X}}
\toprule
\textbf{Benchmark} & \makecell{\textbf{Task-} \\ \textbf{Driven}} & \makecell{\textbf{Real-Rob.} \\ \textbf{Scene}} & \makecell{\textbf{Invari.} \\ \textbf{Eval.}} & \makecell{\textbf{Visual} \\ \textbf{Consist.}} & \makecell{\textbf{Spatial} \\ \textbf{Estim.}} & \makecell{\textbf{Implicit} \\ \textbf{Reasoning}} \\
\midrule
VBench~\cite{huang2024vbench} & \xmark & \xmark & \xmark & \checkmark & \xmark & \xmark \\
WorldModelBench~\cite{li2025worldmodelbench} & \checkmark & $\ocircle$ & \xmark & \xmark & \xmark & \xmark \\
EWMBench~\cite{yue2025ewmbench} & \checkmark & \checkmark & \xmark & \checkmark & \checkmark & \xmark \\
WorldEval~\cite{li2025worldeval} & \checkmark & \checkmark & \xmark & \xmark & \xmark & \xmark \\
WorldScore~\cite{duan2025worldscore} & \xmark & \xmark & $\ocircle$ & \checkmark & \checkmark & \xmark \\
World-in-World~\cite{zhang2025world} & \checkmark & \xmark & \xmark & \checkmark & \xmark & \xmark \\
WorldArena~\cite{shang2026worldarena} & \checkmark & \xmark & $\ocircle$ & \checkmark & \checkmark & \xmark \\
RBench~\cite{deng2026rethinking} & \checkmark & \checkmark & \xmark & \checkmark & \xmark & \xmark \\
\midrule
\textbf{Target-Bench} & \checkmark & \checkmark & \checkmark & \checkmark & \checkmark & \checkmark \\
\bottomrule
\end{tabularx}
\end{table}

\subsubsection{World-Decoder}
\label{sec:world_decoder}
\paragraph{Spatio-temporal Reconstruction.}
As shown in Fig.~\ref{fig:framework}, to extract camera trajectories from \gls{wm}'s generated videos, we employ three state-of-the-art 3D reconstruction methods: \acrshort{vggt}~\cite{wang2025vggtvisualgeometrygrounded}, SpaTracker~\cite{xiao2025spatialtrackerv2}, and ViPE~\cite{huang2025vipe}. Each method processes video frames to reconstruct camera poses.

\smallskip
\textbf{VGGT.} \gls{vggt} is a feed-forward transformer that directly predicts camera poses, depth maps, and point clouds from multi-view images. Given a sequence of $S$ images $\mathcal{I} = \{I_1, I_2, \ldots, I_S\}$, \gls{vggt} first encodes them through a vision transformer to obtain aggregated tokens. A camera head then predicts a pose encoding $\mathbf{p}_s = [\mathbf{T}_s; \mathbf{q}_s; \mathrm{\mathbf{fov}}_s] \in \mathbb{R}^9$ for each frame $I_s$, where $\mathbf{T}_s \in \mathbb{R}^3$ is the camera translation vector, $\mathbf{q}_s \in \mathbb{R}^4$ is the rotation quaternion, and $\mathrm{\mathbf{fov}}_s \in \mathbb{R}^2$ represents the horizontal and vertical field of view. This encoding is converted to standard camera parameters $\mathbf{E}_s = [\mathbf{R}(\mathbf{q}_s) \mid \mathbf{T}_s] \in \mathbb{R}^{3 \times 4}$, where $\mathbf{R}(\mathbf{q}_s)$ converts the quaternion to a rotation matrix.

\smallskip
\textbf{SpaTracker.} SpaTracker extends \gls{vggt} with tracking for better temporal consistency. It adopts a two-stage design: (1) VGGT4Track predicts initial poses $\{\mathbf{E}_s^{(0)}\}_{s=1}^S$ and depths $\{D_s\}_{s=1}^S$; (2) a tracking module refines these poses via point correspondences across frames. Given query points $\mathbf{Q} \in \mathbb{R}^{N \times 3}$ on the first frame, the tracker optimizes camera poses through bundle adjustment:
$\{\mathbf{E}_s^*\} = \arg\min_{\{\mathbf{E}_s\}} \sum_{s=1}^S \sum_{n=1}^N \rho\left(\|\pi(\mathbf{K}_s \mathbf{E}_s \mathbf{X}_n) - \mathbf{x}_{n,s}\|^2\right)$, where $\mathbf{X}_n$ is the 3D position of point $n$, $\pi(\cdot)$ the projection, $\mathbf{x}_{n,s}$ the tracked 2D point, and $\rho(\cdot)$ a robust loss. The result is a refined camera-to-world trajectory $\{\mathbf{C}_{2W,s}\}_{s=1}^S$, where each $\mathbf{C}_{2W,s} \in \mathbb{R}^{4 \times 4}$ is the full camera-to-world transform.

\smallskip
\textbf{ViPE.} ViPE is a \gls{slam}-based visual-inertial pipeline providing metric-scale poses by fusing visual and inertial data. Given RGB frames and \gls{imu} inputs, it performs incremental pose estimation with loop closure. Unlike \gls{vggt} and SpaTracker, ViPE directly outputs paths in metric units via sensor fusion without requiring scale recovery. The final path is a sequence of SE(3) transformations $\{\mathbf{T}_s\}_{s=1}^S$, where each $\mathbf{T}_s \in \mathbb{R}^{4 \times 4}$ encodes rotation and translation in meters.
\paragraph{Scale Factor Recovery.}
Monocular methods such as \gls{vggt} and SpaTracker estimate camera motion only up to an unknown scale. We restore metric consistency at the segment level by anchoring predictions to a single scalar scale factor $\lambda$ derived from real displacement $d_{\mathrm{real}}$ using the first and last ones from the scale factor recovery frames that result in two predicted extrinsic matrices $\mathbf{E}_{f},\mathbf{E}_l \in \mathbb{R}^{3\times 4}$ which are corresponding to the two predicted camera poses enclosed by green dotted box in Fig.~\ref{fig:framework}. We extract the translation vectors $\mathbf{t}_{f}$ and $\mathbf{t}_l$ as the fourth column of $\mathbf{E}_{f}$ and $\mathbf{E}_l$. The predicted displacement is $d_{\mathrm{pred}}=\|\mathbf{t}_{f}-\mathbf{t}_l\|_2$, and the scale factor is $\lambda=d_{\mathrm{real}}/d_{\mathrm{pred}}$. Then for all other predicted future frames, we rescale their predicted translation vectors together, $\mathbf{t}_s^{\mathrm{scaled}}=\lambda\,\mathbf{t}_s$ for $s=1,\ldots,S$ with $S$ be the total number of all frames. This lightweight mechanism preserves relative geometry while lifting trajectories to metric values, enabling quantitative comparison.

\subsubsection{Path Evaluation}
\label{subsubsec:eval_metrics}

\paragraph{Metrics.} We evaluate the planning results of video world models 
using a comprehensive suite of metrics that assess accuracy, goal-reaching capability, and path consistency. All metrics assume 2D trajectories represented as sequences of positions, with ground truth $\mathbf{s}^{\mathrm{GT}} = \{s_1, s_2, \ldots, s_T\}$ and prediction $\hat{\mathbf{s}} = \{\hat{s}_1, \hat{s}_2, \ldots, \hat{s}_T\}$, where $s_t, \hat{s}_t \in \mathbb{R}^2$ and $T$ is the number of time steps.

\begin{figure}[t]
\centering
\includegraphics[width=0.9\columnwidth]{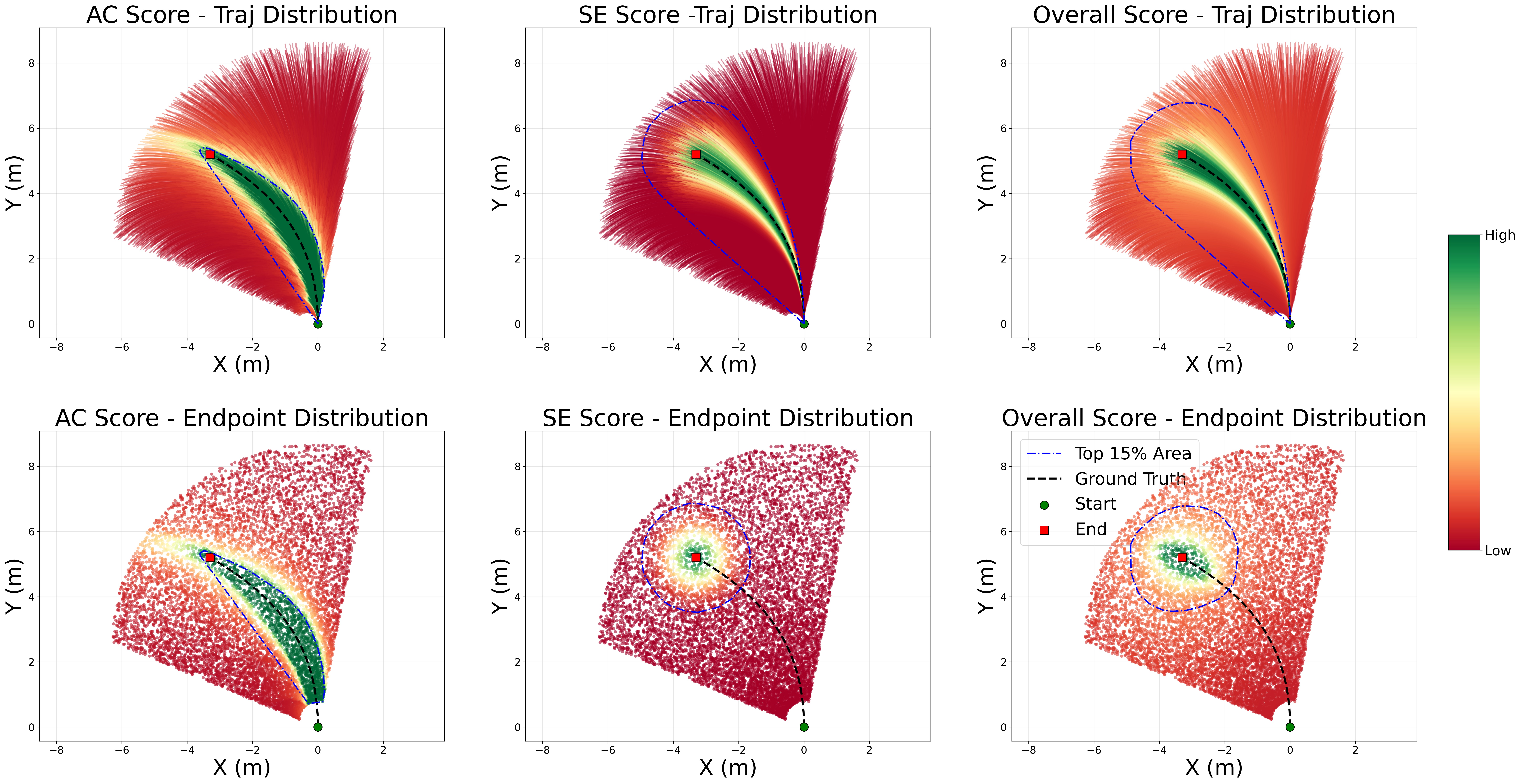}
\caption{Illustration of metrics AC and SE with trajectory and endpoint score distribution heatmaps. Plots visualize scores over a fixed set of candidate trajectories. Trajectories with good Approach Consistency can have a bad Endpoint (Far away from the target). Trajectories with a good Endpoint can have a bad Approach Consistency (Not within the oval corridor). Trajectories with a good Overall Score have both good Approach Consistency and Endpoint.}
\label{fig:heatmaps}
\end{figure}  

\paragraph{\gls{ade}.}
Measures the average L2 distance between predicted and ground truth positions across all timesteps:
$
\mathrm{ADE} = \frac{1}{T} \sum_{t=1}^{T} \|\hat{s}_t - s_t\|_2
$.

\paragraph{\gls{fde}.}
Evaluates the distance between final positions:
$\mathrm{FDE} = \|\hat{s}_T - s_T\|_2$.

\paragraph{\gls{mr}.}
Computes the percentage of predicted points that deviate beyond a threshold $\tau$ (default: $2.0$ m):
$
\mathrm{MR} = \frac{1}{T} \sum_{t=1}^{T} \mathbb{I}\left[\|\hat{s}_t - s_t\|_2 > \tau\right] \cdot 100
$
\paragraph{\gls{se}.}
Uses a Gaussian penalty to measure the proximity of the endpoint to the target:
\begin{equation}
\mathrm{SE} = \exp\left(-\frac{\|\hat{s}_T - s_T\|_2^2}{2\sigma^2}\right)
\end{equation}
where $\sigma = 0.6$ m is the tolerance. $\mathrm{SE} \in [0,1]$ where $1$ indicates perfect alignment.

\paragraph{\gls{ac}.}
Evaluates whether the predicted trajectory stays within a progress-dependent corridor around the ground truth path. We uniformly sample $M=20$ reference points along the ground truth trajectory and assign the corridor radius at the $i$-th reference point as $\sigma_i$:
\begin{equation}
\sigma_i = \sigma_{\min} + (\sigma_{\max} - \sigma_{\min}) \exp\left(-\frac{(p_i - 0.5)^2}{2\beta^2}\right)
\end{equation}
where $p_i = i/(M-1)$ is the normalized progress, $\sigma_{\min}=0.15$ m and $\sigma_{\max}=0.5$ m are the minimum and maximum corridor radius, and $\beta=0.25$. A predicted point $\hat{s}_j$ is covered if $\min_i \|\hat{s}_j - s_i^{\mathrm{GT}}\|_2 \leq \sigma_i$. The \gls{ac} score is given as:
\begin{equation}
\mathrm{AC} = 
\begin{cases}
1, & N_{\mathrm{c}} = N_{\mathrm{p}} \\[6pt]
\exp\left(-\gamma \cdot \dfrac{N_{\mathrm{p}} - N_{\mathrm{c}}}{N_{\mathrm{p}}}\right), & \text{otherwise}
\end{cases}
\end{equation}
where $N_{\mathrm{p}}$ is the total number of predicted points, $N_{\mathrm{c}}$ is the number covered by the corridor, and $\gamma=5$. Importantly, this metric design supports non-uniqueness of valid path. The ground truth path is treated as a rough reference rather than a single correct solution, conditioned on the initial heading. As shown in Fig.~\ref{fig:heatmaps}, various trajectories with similar motional tendency can achieve the same level of high score, which means the score is tendency based, instead of trajectory based.

\paragraph{\gls{wo} Score.}
Aggregates all metrics into a unified score $\in [0,1]$ (the higher the better):
\begin{equation}
\begin{aligned}
S_{\mathrm{overall}} &= w_{\mathrm{ADE}} \cdot \exp\left(-\frac{\mathrm{ADE}}{\tau_{\mathrm{ADE}}}\right) + w_{\mathrm{FDE}} \cdot \exp\left(-\frac{\mathrm{FDE}}{\tau_{\mathrm{FDE}}}\right) \\
&\quad + w_{\mathrm{MR}} \cdot \left(1 - \frac{\mathrm{MR}}{100}\right) + w_{\mathrm{SE,AC}} \cdot \mathrm{SE} \cdot \mathrm{AC}
\end{aligned}
\end{equation}
with default weights: $w_{\mathrm{ADE}}=0.075$, $w_{\mathrm{FDE}}=0.125$, $w_{\mathrm{MR}}=0.125$, $w_{\mathrm{SE,AC}}=0.675$, and scale parameters $\tau_{\mathrm{ADE}}=\tau_{\mathrm{FDE}}=1.0$ m. The weight design prioritizes task success; therefore, a dominant weight is assigned to $\mathrm{SE}\!\cdot\!\mathrm{AC}$ to emphasize endpoint proximity and approach consistency, while keeping $\mathrm{ADE}$, $\mathrm{FDE}$, and $\mathrm{MR}$ lightweight to avoid over-emphasizing point-wise trajectory overlapping.

%% file: sec/4_exp.tex
\section{Experiments}

\begin{table*}[h]
\caption{Evaluation results with \gls{vggt}.
\textit{\footnotesize $^\dagger$ Image-to-Video (I2V) models, $^\S$ fine-tuned, $^\P$ fine-tuned with data augmentation.}}
\label{tab:vggt_metric}
\centering
\setlength{\tabcolsep}{3pt}
\renewcommand{\arraystretch}{1.1}
\small
\resizebox{\textwidth}{!}{%
\begin{tabular}{cccc|ccccc|c}
\toprule
\multicolumn{1}{c}{Spatial} & 
\multirow{2}{*}{World Model} & \multicolumn{1}{c}{Open} &
\multicolumn{1}{c|}{Video Length (s)} & 
\multicolumn{5}{c|}{Metrics} & 
\multirow{2}{*}{WO~$\uparrow$}\\
\multicolumn{1}{c}{Method}& & \multicolumn{1}{c}{Source} &\multicolumn{1}{c|}{\& Resolution} & \multicolumn{1}{c}{FDE~$\downarrow$} & \multicolumn{1}{c}{ADE~$\downarrow$} & \multicolumn{1}{c}{MR~$\downarrow$} & \multicolumn{1}{c}{SE~$\uparrow$} & \multicolumn{1}{c|}{AC~$\uparrow$}  &   \\
\midrule
\multirow{12}{*}{VGGT~\cite{wang2025vggtvisualgeometrygrounded}} 
    & \fadedtext{gt\_video} & \fadedtext{-} & \fadedtext{8 - 720p} & \fadedtext{0.203} & \fadedtext{0.580} & \fadedtext{11.08} & \fadedtext{0.901} & \fadedtext{0.993} & \fadedtext{0.862} \\
    & Sora 2 & \xmark  & 8 - 720p & 1.912 & 1.289 & 54.84 & 0.160 & 0.788 & 0.207 \\
    & Veo 3.1 & \xmark  & 8 - 720p & 3.125 & 2.432 & 77.25 & 0.212 & 0.450 & 0.210 \\
    & Veo 3.1-fast & \xmark  & 8 - 720p  & 2.863 & 1.798 & 57.42 & 0.140 & 0.691 & 0.192 \\
    & Wan2.5-Preview$^\dagger$ & \xmark & 10 - 720p & 2.478 & 1.596 & 61.69 & 0.182 & 0.873 & 0.215 \\
    & Wan2.2-Plus$^\dagger$ & \xmark & 5 - 1080p & 1.377 & 1.044 & 45.25 & 0.240 & 0.686 & 0.290 \\
    & Wan2.2-Flash$^\dagger$ & \xmark & 5 - 720p & 1.362 & 1.005 & 38.75 & 0.292 & 0.746 & 0.341 \\
    & Wan2.1-Plus$^\dagger$ & \xmark & 5 - 720p & 2.782 & 1.594 & 53.90 & 0.090 & 0.715 & 0.157 \\
    & Wan2.1-Turbo$^\dagger$ & \xmark & 5 - 720p & 4.243 & 1.850 & 64.21 & 0.000 & \textbf{1.000} & 0.059 \\
    & Wan2.2-5B-base & \cmark & 8 - 720p   & 3.666 & 1.636 & 58.11 & 0.012 & 0.944 & 0.084 \\
    & Wan2.2-5B-FT$^\S$ & \cmark & 8 - 720p   & 1.320 & 0.897 & 31.91 & 0.261 & 0.787 & 0.330 \\
    & Wan2.2-5B-FT-DA$^\P$ & \cmark & 8 - 720p   & \textbf{1.050} & \textbf{0.816} & \textbf{26.71} & \textbf{0.333} & 0.810 & \textbf{0.394} \\
\bottomrule
\end{tabular}%
}
\end{table*}

\begin{figure}[t]
\centering
\includegraphics[width=1\columnwidth]{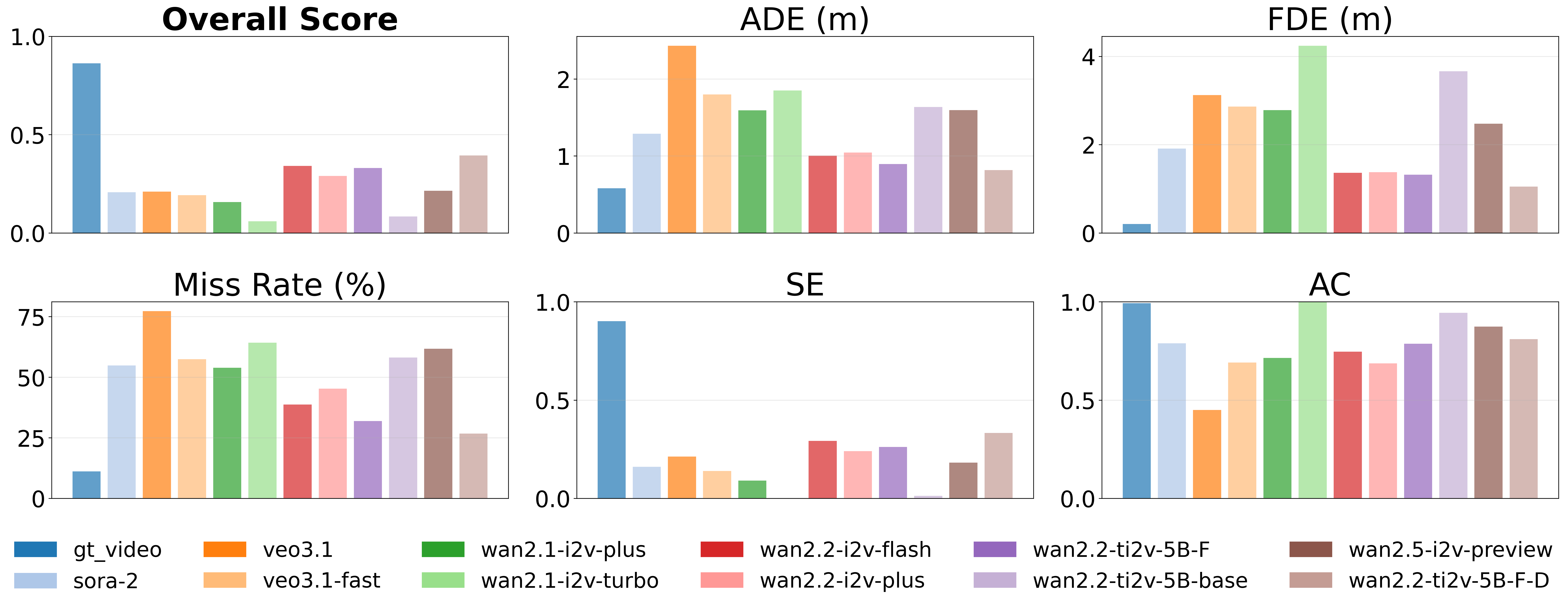}
\caption{World model performance comparison with VGGT as world-decoder's spatio-temporal reconstruction tool.}
\label{fig:model_compare_vggt}
\end{figure}   

Our proposed evaluation framework uses five key metrics detailed in Sec.~\ref{subsubsec:eval_metrics}. The analysis includes several state-of-the-art world models, including Sora 2~\cite{sora2}, Veo 3.1~\cite{veo3deepmind}, Veo 3.1-fast, and multiple variants of the Wan series (Wan2.5-I2V-Preview, Wan2.2-Plus, Wan2.2-Flash, Wan2.1-Plus, and Wan2.1-Turbo)~\cite{wan2025wan}. In addition, we evaluate the fine-tuned Wan2.2-TI2V-5B models.

\subsection{Implementation Details}
For model fine-tuning we use the LoRA training framework offered by DiffSynth-Studio~\cite{DiffSynthStudio_wanvideo}, with 8× NVIDIA A800 80 GB GPUs. For inference of the fine-tuned models on the Target Benchmark we utilize one single NVIDIA RTX PRO 6000 Blackwell 96 GB GPU, while closed-source models are accessed directly through their official APIs. All generated videos are produced at 720p or 1080p resolution with durations ranging from 5 to 10 seconds. All evaluation experiments are conducted on a workstation equipped with an NVIDIA RTX 4090 GPU.

\subsection{Evaluation for off-the-shelf Models}
Table~\ref{tab:vggt_metric} shows the evaluation results using \gls{vggt} as the spatio-temporal reconstruction tool. Among all evaluated off-the-shelf models, Wan2.2-Flash achieves the best overall performance with a weighted overall score of 0.341. Specifically, it obtains the lowest errors in \gls{fde} (1.362m), \gls{ade} (1.005m), and \gls{mr} (38.75\%), while achieving the highest \gls{se} (0.292). 
Figure~\ref{fig:model_compare_vggt} visualizes the performance comparison across all models. 

\paragraph{Displacement Errors:} \gls{ade} values range from 1.0m to 2.4m across different models. Veo 3.1 exhibits the highest displacement error (2.432m), while Wan2.2-Flash and Wan2.2-Plus achieve the best accuracy (around 1.0m). For \gls{fde}, errors show wider variation, with Wan2.1-Turbo displaying the highest \gls{fde}.

\paragraph{Reliability Metrics:} The MR varies significantly from 38.75\% to 77.25\%. Veo 3.1 shows the highest MR (77.25\%), indicating poor trajectory quality, while Wan2.2-Flash achieves the best (38.75\%). Most models score below 0.3 on the SE metric, suggesting challenges in reaching target endpoints accurately.

\paragraph{Consistency:} Wan2.1-Turbo achieves the highest Approach Consistency (1.000), indicating perfect directional alignment, while Veo 3.1 shows the poorest performance (0.450). Sora 2 demonstrates good consistency at 0.788.

\paragraph{Explicit vs. Implicit Targets:} As shown in Table~\ref{tab:implicit_metric}, performance on implicitly defined targets closely matches that on explicit targets, with small, model-specific fluctuations. This indicates that current video \glspl{wm} can understand semantic goals even when they are not explicitly defined.

\subsection{Fine-tuned Models}
\label{subsec:fine_tuned_models}
We use the train split from our dataset for fine-tuning the open-sourced Wan2.2-TI2V-5B model. The text prompt and the first video frame are taken as inputs. We evenly sample 121 frames from each video to form a ground truth frame sequence. The model is fine-tuned under two settings: without and with data augmentation. For data augmentation, we apply a shifting-frame strategy: for each video of a scenario, we generate four clips of 121 frames starting from different offsets, resulting in a four times large training set while preserving similar temporal coverage and reusing the same captions of each sample. Table~\ref{tab:vggt_metric} evaluates all models on unseen data using \gls{vggt} for path reconstruction. The fine-tuned Wan2.2-5B (\texttt{Wan2.2-5B-FT}) improves its score from 0.084 to 0.330.
The augmented version (\texttt{Wan2.2-5B-FT-DA}) outperforms the base model by more than 469\% and achieves the best overall score.

\begin{table}[t]
\caption{Evaluation results with explicit and implicit semantic targets.}
\label{tab:implicit_metric}
\vspace{-1.7mm}
\begin{center}
\setlength{\tabcolsep}{3pt}
\renewcommand{\arraystretch}{1.25}
\small
\resizebox{0.65\linewidth}{!}{
\begin{tabular}{cc|ccccc|c}
\toprule
\multirow{2}{*}{Data} & 
\multicolumn{1}{c}{World} &
\multicolumn{5}{|c|}{Metrics} & 
\multirow{2}{*}{WO~$\uparrow$}\\
& \multicolumn{1}{c}{Model} & \multicolumn{1}{|c}{FDE~$\downarrow$} & \multicolumn{1}{c}{ADE~$\downarrow$} & \multicolumn{1}{c}{MR~$\downarrow$} & \multicolumn{1}{c}{SE~$\uparrow$} & \multicolumn{1}{c|}{AC~$\uparrow$}  &   \\
\midrule
\multirow{7}{*}{Explicit Target}
    & \fadedtext{gt\_video} & \fadedtext{0.210} & \fadedtext{0.586} & \fadedtext{10.71} & \fadedtext{0.892} & \fadedtext{0.994} & \fadedtext{0.857} \\
    & Sora 2 & 1.979 & 1.381 & 56.93 & 0.165 & 0.740 & 0.206 \\
    & Veo 3.1 & 3.275 & 2.598 & 80.69 & 0.172 & 0.419 & 0.175 \\
    & Veo 3.1-fast & 3.225 & 1.928 & 57.37 & 0.136 & 0.685 & 0.188 \\
    & Wan2.5-I2V-Preview & 2.449 & 1.656 & 61.70 & 0.182 & \textbf{0.826} & 0.216 \\
    & Wan2.2-I2V-Plus & \textbf{1.413} & 1.028 & 42.25 & 0.242 & 0.724 & 0.298 \\
    & Wan2.2-I2V-Flash & 1.418 & \textbf{1.006} & \textbf{35.69} & \textbf{0.295} & 0.776 & \textbf{0.349} \\
\midrule
\multirow{7}{*}{Implicit Target}
    & \fadedtext{gt\_video} & \fadedtext{0.193} & \fadedtext{0.573} & \fadedtext{11.61} & \fadedtext{0.913} & \fadedtext{0.992} & \fadedtext{0.871} \\
    & Sora 2 & 1.819 & 1.160 & 51.91 & 0.154 & 0.856 & 0.210 \\
    & Veo 3.1 & 2.914 & 2.200 & 72.43 & 0.269 & 0.493 & 0.259 \\
    & Veo 3.1-fast & 2.355 & 1.616 & 57.49 & 0.144 & 0.700 & 0.198 \\
    & Wan2.5-I2V-Preview & 2.519 & 1.512 & 61.67 & 0.181 & \textbf{0.941} & 0.214 \\
    & Wan2.2-I2V-Plus & 1.327 & 1.067 & 49.46 & 0.238 & 0.634 & 0.278 \\
    & Wan2.2-I2V-Flash & \textbf{1.283} & \textbf{1.002} & \textbf{43.05} & \textbf{0.288} & 0.706 & \textbf{0.331} \\
\bottomrule
\end{tabular}
}
\end{center}
\end{table}

\begin{table*}[t]
\caption{Evaluation results of SpaTracker and ViPE.
\textit{\footnotesize  $^\dagger$ I2V models, $^\S$ Wan2.2-TI2V-5B fine-tuned model, $^\P$ Wan2.2-TI2V-5B model fine-tuned with data augmentation.}}
\label{tab:reconstruction_metric}
\centering
\setlength{\tabcolsep}{3pt}
\renewcommand{\arraystretch}{1.25}
\small
\resizebox{\textwidth}{!}{%
\begin{tabular}{c|cccccc|cccccc}
\toprule
\multirow{2}{*}{World Model} &
\multicolumn{6}{c|}{SpaTracker~\cite{xiao2025spatialtrackerv2}} &
\multicolumn{6}{c}{ViPE~\cite{huang2025vipe}} \\
& FDE$\downarrow$ & ADE$\downarrow$ & MR$\downarrow$ & SE$\uparrow$ & AC$\uparrow$ & WO$\uparrow$
& FDE$\downarrow$ & ADE$\downarrow$ & MR$\downarrow$ & SE$\uparrow$ & AC$\uparrow$ & WO$\uparrow$ \\
\midrule
    \fadedtext{gt\_video} & \fadedtext{0.526} & \fadedtext{0.600} & \fadedtext{17.19} & \fadedtext{0.59} & \fadedtext{0.95} & \fadedtext{0.623} & \fadedtext{1.345} & \fadedtext{0.845} & \fadedtext{30.22} & \fadedtext{0.19} & \fadedtext{0.89} & \fadedtext{0.291}\\
    Sora 2 & 1.913 & 1.257 & 51.33 & 0.14 & 0.82 & 0.208 & 2.147 & 1.245 & 51.04 & 0.09 & 0.85 & 0.170\\
    Veo 3.1 & 2.607 & 2.133 & 78.49 & 0.20 & 0.42 & 0.194 & 2.916 & 1.945 & 55.07 & 0.16 & 0.62 & 0.218 \\
    Veo 3.1-fast & 3.402 & 2.150 & 62.89 & 0.15 & 0.59 & 0.191 & 2.601 & 1.715 & 58.76 & 0.15 & 0.62 & 0.202\\
    Wan2.5-Preview$^\dagger$ & 2.379 & 1.510 & 61.93 & 0.18 & 0.88 & 0.213 & 2.686 & 1.608 & 59.52 & 0.10 & 0.83 & 0.156\\
    Wan2.2-Plus$^\dagger$ & 1.409 & 1.631 & 73.19 & 0.25 & 0.68 & 0.253 & \textbf{1.621} & \textbf{0.976} & \textbf{36.65} & \textbf{0.18} & 0.80 & \textbf{0.265}\\
    Wan2.2-Flash$^\dagger$ & 1.407 & 1.572 & 71.06 & 0.28 & 0.71 & 0.284 & 1.687 & 1.006 & 37.42 & 0.17 & 0.81 & 0.260\\
    Wan2.1-Plus$^\dagger$ & 2.810 & 2.110 & 63.08 & 0.08 & 0.67 & 0.141 & 2.858 & 1.510 & 53.70 & 0.04 & 0.76 & 0.125 \\
    Wan2.1-Turbo$^\dagger$ & 4.219 & 1.833 & 63.45 & 0.00 & \textbf{1.00} & 0.061 & 4.256 & 1.856 & 64.35 & 0.00 & \textbf{1.00} & 0.059\\
    Wan2.2-TI2V-5B & 3.661 & 1.666 & 57.05 & 0.03 & 0.89 & 0.104 & 3.773 & 1.705 & 60.27 & 0.01 & 0.91 & 0.083\\
    Wan2.2-5B-FT$^\S$ & 1.342 & 0.951 & 38.53 & 0.23 & 0.74 & 0.300 & 1.980 & 1.079 & 42.76 & 0.12 & 0.80 & 0.208\\
    Wan2.2-5B-FT-DA$^\P$ & \textbf{1.055} & \textbf{0.875} & \textbf{33.99} & \textbf{0.33} & 0.75 & \textbf{0.385} & 1.841 & 1.023 & 39.01 & 0.11 & 0.83 & 0.214\\
\bottomrule
\end{tabular}%
}
\end{table*}

\subsection{Ablation Study}
\begin{figure}[t]
\centering
\includegraphics[width=1\columnwidth]{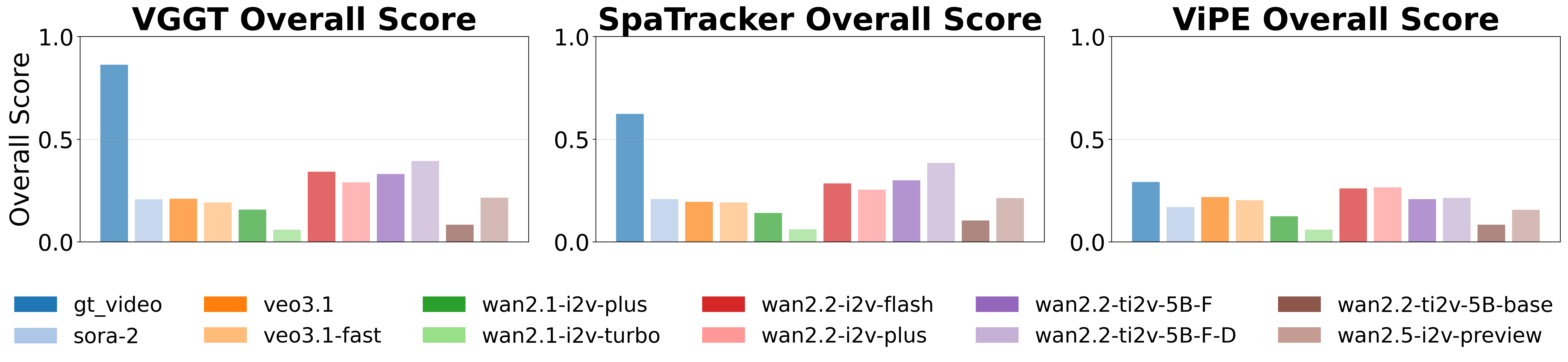}
\caption{Overall score comparison between different spatio-temporal reconstruction tools. Detailed evaluation results can be found in the \textbf{appendix}.}
\label{fig:overall_score_comparison}
\end{figure}

\paragraph{Comparison Among Reconstruction Tools.} Fig.~\ref{fig:overall_score_comparison}, Table~\ref{tab:vggt_metric} and Table~\ref{tab:reconstruction_metric} compare the performance of three spatio-temporal reconstruction tools: \gls{vggt}, SpaTracker, and ViPE. \gls{vggt} achieves the best weighted overall score of 0.862 with the ground truth videos. SpaTracker shows lower performance, while ViPE produces the weakest results. The high score achieved by \gls{vggt} on ground truth videos confirms that decoded trajectories align well with ground truth trajectories, validating our evaluation approach. Notably, this component is replaceable in our framework as temporal reconstruction methods continue to advance.

\begin{table}[h]
\caption{Evaluation results with different path planning horizons, with \gls{vggt}.}
\label{tab:world_metric}
\begin{center}
\setlength{\tabcolsep}{3pt}
\renewcommand{\arraystretch}{1.20}
\resizebox{0.6\linewidth}{!}{
\begin{tabular}{cc|ccccc|c}
\toprule
\multirow{2}{*}{Model} & 
\multirow{2}{*}{Horizon} &
\multicolumn{5}{c|}{Metrics} & 
\multirow{2}{*}{WO~$\uparrow$}\\
& & \multicolumn{1}{c}{FDE~$\downarrow$} & \multicolumn{1}{c}{ADE~$\downarrow$} & \multicolumn{1}{c}{MR~$\downarrow$} & \multicolumn{1}{c}{SE~$\uparrow$} & \multicolumn{1}{c|}{AC~$\uparrow$}  &   \\
\midrule
\multirow{3}{*}{Wan2.2-I2V-Flash}
    & 8s & 1.362 & 1.005 & 38.75 & 0.292 & 0.746 & 0.341 \\
    & 6s & 1.393 & 0.762 & 23.76 & 0.261 & 0.579 & 0.337 \\
    & 4s & 1.278 & 0.713 & 23.18 & 0.290 & 0.574 & 0.363 \\
\midrule
\multirow{3}{*}{Wan2.2-I2V-Plus}
    & 8s & 1.377 & 1.044 & 45.25 & 0.240 & 0.686 & 0.290 \\
    & 6s & 1.390 & 0.789 & 25.36 & 0.263 & 0.584 & 0.338 \\
    & 4s & 1.400 & 0.783 & 25.91 & 0.265 & 0.534 & 0.339 \\
\bottomrule
\end{tabular}
}
\end{center}
\end{table}

\subsection{Sensitivity to the Planning Horizon Length}
To assess how the path planning horizon influences \gls{wm} performance, Table \ref{tab:world_metric} reports results for two \glspl{wm} evaluated at three horizons: 8 s, 6 s, and 4 s.
Reducing the horizon from 6 s to 4 s yields an 8\% \gls{wo} improvement for \texttt{Wan2.2-I2V-Flash}, while shortening it from 8 s to 6 s increases the weighted score of \texttt{Wan2.2-I2V-Plus} by more than 17\%.
Overall, the weighted score consistently improves as the horizon decreases, suggesting that \glspl{wm} are more reliable when planning on shorter temporal windows.

%% file: sec/5_ds.tex
\section{Discussion}

\paragraph{World Models Performance.}
The best off-the-shelf video \gls{wm}, \texttt{Wan2.2-Flash}, achieves only 0.341 overall score. This indicates a significant gap between current world model capabilities and reliable path planning. However, qualitative inspection of generated videos reveals that most models correctly understand the semantic target and show plausible motion tendencies. Additionally, our benchmark includes scenarios with implicit semantic targets, which makes our benchmark challenging for video world models to harvest good scores. 

\paragraph{Holistic Evaluation: Beyond Visual Quality.}
Target-Bench intrinsically evaluates three capabilities simultaneously: (1) spatio-temporal consistency, (2) semantic reasoning, and (3) geometric path planning accuracy.
Blurred frames, temporal discontinuities, or spatial warping will directly result in poor 3D reconstruction which causes low path accuracy. The holistic assessment of our work fundamentally differs from existing benchmarks that isolate individual dimensions (visual quality, physics consistency) and miss their integration. 


\paragraph{Real-World Navigation.} Given an observation and semantic goal, \glspl{wm} predict future frames depicting motion toward the target, and our world-decoder extracts a path for the robot to execute. This has potential for future use in closed-loop robot navigation in unstructured, mapless environments. Recent work~\cite{ball2025genie3, yu2025context} shows \glspl{wm} can explore new spaces while retaining information in latent memory. Combined with our decoder, robots could follow semantic goals using only visual predictions. As shown in the \textbf{appendix}, we also show some preliminary results in terms of deployment of video world model with world-decoder on a quadruped robot regarding real-world navigation tasks.

%% file: sec/6_ccl.tex
\section{Conclusion and Future Work}
We introduced Target-Bench for evaluating video world models on mapless path planning toward semantic targets. Our evaluation pipeline recovers planned paths from generated videos and measures planning performance against SLAM-based ground-truth robot paths. The best off-the-shelf video \gls{wm} achieves a relatively low score. However, fine-tuning an open-source 5B-parameter model on only 325 scenarios from our dataset surpasses the best off-the-shelf video \glspl{wm}. This suggests that \glspl{wm} can effectively learn navigation tasks from limited real-world data, showing promising potential for robot path planning. Future work could extend the current framework towards closed-loop re-planning on the robot, and studying how latent memory supports complex receding-horizon navigation tasks. Better reconstruction tools are expected to further improve path planning accuracy, facilitating real-world deployment of \glspl{wm} in robotics.